\documentclass[conference]{IEEEtran}
\IEEEoverridecommandlockouts
\usepackage{cite}
\usepackage{amsmath,amssymb,amsfonts}
\usepackage{algorithmic}
\usepackage{graphicx}
\usepackage{textcomp}
\usepackage{xcolor}
\usepackage{enumitem}
\usepackage{booktabs}
\usepackage{multirow}

\def\BibTeX{{\rm B\kern-.05em{\sc i\kern-.025em b}\kern-.08em
    T\kern-.1667em\lower.7ex\hbox{E}\kern-.125emX}}
\begin{document}

\title{Large Language Models Meet Contrastive Learning: Zero-Shot Emotion Recognition Across Languages}

\author{Heqing Zou\textsuperscript{\ddag}, 
 Fengmao Lv\textsuperscript{$\flat$}, 
 Desheng Zheng\textsuperscript{\S}\textsuperscript{\pounds}, 
Eng Siong Chng\textsuperscript{\ddag}, Deepu Rajan\textsuperscript{\ddag}
 \\
\textsuperscript{\ddag}Nanyang Technological University \textsuperscript{$\flat$}Southwest Jiaotong University  \\ \textsuperscript{\S}Kash Institute of Electronics and Information Industry \textsuperscript{\pounds}Southwest Petroleum University \\
  }

\maketitle

\begin{abstract}
Multilingual speech emotion recognition aims to estimate a speaker's emotional state using a contactless method across different languages. However, variability in voice characteristics and linguistic diversity poses significant challenges for zero-shot speech emotion recognition, especially with multilingual datasets. In this paper, we propose leveraging contrastive learning to refine multilingual speech features and extend large language models for zero-shot multilingual speech emotion estimation. Specifically, we employ a novel two-stage training framework to align speech signals with linguistic features in the emotional space, capturing both emotion-aware and language-agnostic speech representations. To advance research in this field, we introduce a large-scale synthetic multilingual speech emotion dataset, M5SER. Our experiments demonstrate the effectiveness of the proposed method in both speech emotion recognition and zero-shot multilingual speech emotion recognition, including previously unseen datasets and languages. Our introduced dataset and related code will be available on GitHub \footnote{https://github.com/Vincent-ZHQ/MSER}.
\end{abstract}


\section{Introduction}
Unlike monolingual speech emotion recognition (SER) \cite{busso2008iemocap, livingstone2018ryerson, chen2024vesper}, which focuses on a single language, multilingual speech emotion recognition (MSER) \cite{al2022transformer} leverages multiple languages, making it suitable for a wide range of multilingual human-computer interaction systems. A common approach to MSER involves training and testing a classifier on a specific human-annotated corpus \cite{zehra2021cross, sharma2022multi}. However, these frameworks are often tailored to the characteristics of the training datasets, including languages, speakers, and emotion categories, limiting their generalizability to other datasets \cite{ma2024emobox}. Zero-shot MSER, with its ability to estimate emotions on unseen datasets and languages, is crucial for direct in-the-wild applications, especially for languages with limited audio training resources \cite{li2023zero}.

In recent years, cross-corpus SER \cite{gao2023adversarial} has been explored to assess its performance across multiple datasets. Some methods train SER models on a specific dataset and evaluate their performance on other datasets with the same language, while others assess datasets in different languages. Although performance significantly decreases during cross-corpus evaluations \cite{lian2024exploring}, these approaches provide valuable insights into learning voice-independent, emotion-aware speech representations \cite{zhao2024emotion}. Multilingual SER aims to estimate a speaker's emotions across multiple corpora in different languages. These multilingual systems can typically be applied directly to various speech resources using the same model. However, their performance often falls short compared to corresponding monolingual recognition methods. Additionally, MSER methods still struggle with limited generalization to unseen languages \cite{li2023zero}. 

The challenges of zero-shot MSER are complex and multifaceted. Individual voice variability among speakers can significantly impact emotion interpretation, as distinct vocal characteristics convey emotional cues differently \cite{gomez2024speech}. Additionally, the diversity of languages adds complexity, as emotional expressions are influenced by linguistic and cultural contexts, leading to variations in articulation. The scarcity of data for minority languages exacerbates these challenges, limiting the model's ability to learn robust emotion representations \cite{ma2024emobox}.

Recent advancements in audio-conditioned large language models (AcLLMs), such as QWen-Audio \cite{chu2023qwen}, have demonstrated exceptional multilingual speech reasoning capabilities by integrating continuous speech representations into large language models (LLMs). Most AcLLMs focus on general speech understanding, either performing audio analysis or generating textual responses. These models typically leverage speech foundation models, like Whisper \cite{radford2023robust}, to capture speech features and train on extensive audio-text pairs. Some other methods extend AcLLMs for emotion estimation \cite{santoso2024large}. For instance, SECap \cite{xu2024secap} introduces a novel speech emotion captioning framework that effectively describes speech emotions using natural language, while SELM \cite{bukhari2024selm} incorporates automatic speech recognition (ASR) to enhance SER performance. Despite these advances, these methods remain limited to end-to-end fine-tuning on specific target datasets, and they lack the capability to perform zero-shot emotion estimation across diverse datasets or languages.

In this paper, we propose a two-stage framework to contrastively align multilingual speech representations with the target language space, capturing emotion-aware and language-agnostic speech information for zero-shot multilingual speech emotion recognition (MSER). Our framework comprises a multilingual audio encoder, a speech connector, and a large language model (LLM). We reformulate the speech emotion classification task as an emotion word prediction task, leveraging the LLM's zero-shot reasoning capabilities. In our two-stage training pipeline, we first contrastively train the speech emotion decoding task using public, cleaner English speech datasets, aligning emotion-related language information with emotion-aware speech representations. To address the scarcity of MSER datasets, we introduce M5SER, a large-scale synthetic speech emotion corpus generated by emotion-preserving speech foundation models with self-annotated emotion labels. In the second stage, we refine the model through contrastive fine-tuning with the introduced multilingual sources, enabling it to capture language-agnostic, emotion-aware speech information for zero-shot MSER from synthetic data. Our method achieves state-of-the-art performance in both traditional SER and zero-shot MSER evaluations on datasets with previously unseen languages. In summary, our contributions are:

\begin{itemize}[leftmargin=*]
\setlength\itemsep{0em} 
    \item To facilitate zero-shot MSER, we propose a novel training framework that contrastively aligns multilingual speech information with shared emotion-aware language features. 
    \item We introduce a large-scale MSER dataset, M5SER, generated using emotion-preserving speech foundation models to advance research on multilingual speech emotion estimation. 
    \item Extensive experiments on traditional SER and zero-shot MSER benchmarks demonstrate the effectiveness of our proposed method.
\end{itemize}

\section{Related Works}

\subsection{Speech Emotion Recognition (SER)} 
SER systems estimate a speaker's emotional state from audio. Traditional SER follows an end-to-end approach on a single dataset \cite{zou2022speech, chen2024vesper}. Beyond this, cross-dataset SER \cite{gao2023adversarial} trains on one or multiple datasets and evaluates on different ones, often facing performance drops due to distribution shifts. Multilingual SER \cite{sharma2022multi, ma2024emobox} extends SER to multiple languages within the same datasets. Cross-lingual SER \cite{li2023zero} tackles both dataset and language shifts, requiring models to learn language-agnostic speech representations for robust emotional understanding across languages.

\subsection{Speech Large Language Models}
The evolution from LLMs to Multimodal Large Language Models (MM-LLMs) has revolutionized speech and audio processing. Early LLMs like GPT-3 \cite{floridi2020gpt} were text-only, whereas MM-LLMs integrate speech and text \cite{chu2023qwen}, leveraging connectors to align audio and textual features \cite{radford2023robust, xu2024secap}. This enables advanced tasks such as emotion recognition, audio reasoning, and cross-modal understanding. In the realm of SER, this multimodal approach allows for deeper contextual comprehension, as MM-LLMs can now jointly analyze verbal and acoustic signals, capturing emotional nuances that were previously missed by text-only models.

\subsection{Contrastive learning}
Contrastive learning has emerged as a powerful self-supervised technique for representation learning \cite{ericsson2022self, zou2023unis}. It captures informative features by bringing similar data points closer and pushing dissimilar ones apart, reducing reliance on labeled data. The core idea is to bring positive sample pairs (e.g., different augmentations of the same data) closer in the feature space while pushing negative pairs (e.g., unrelated data points) further apart. Early successes in models like SimCLR \cite{chen2020simple} revolutionized image and vision tasks by effectively learning from large unlabeled datasets. This approach has been extended to various modalities, including speech, text, and multimodal data \cite{zou2024cross}. Recent advances use contrastive learning for speech emotion recognition, aligning emotion-aware speech features across languages and datasets. This improves generalization to unseen tasks, including zero-shot learning, making it vital for modern representation learning. 

\begin{figure}[htbp]
\vspace{-0.5cm}
\centerline{\includegraphics[width=1.0\linewidth]{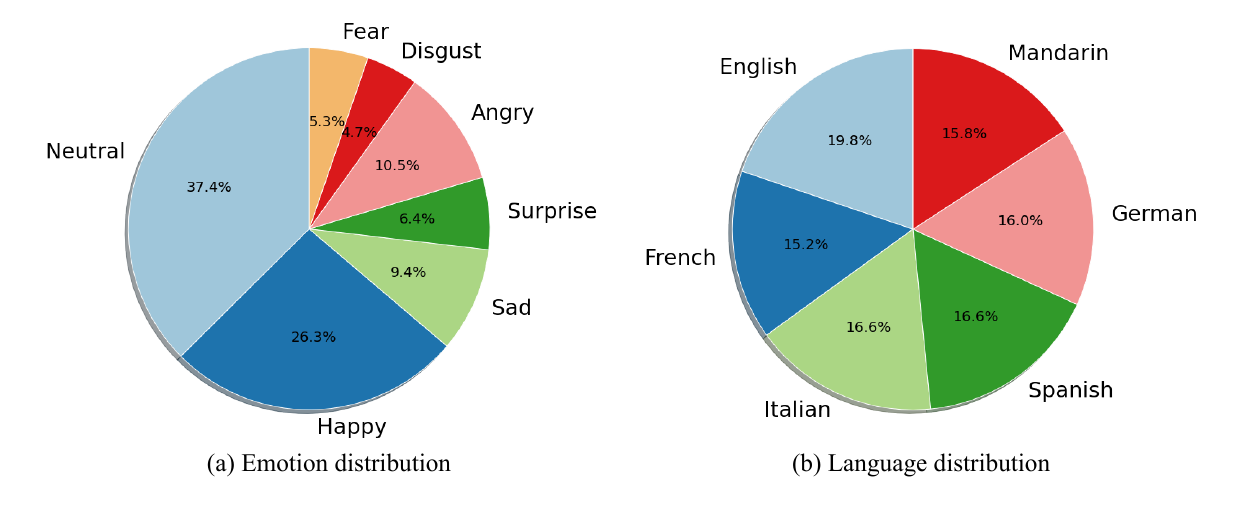}}
\vspace{-0.3cm}
\caption{M5SER: (a) Emotion distribution, (b) Language distribution.}
\label{fig02}
\vspace{-0.3cm}
\end{figure}

\begin{figure*}[htbp]
\centerline{\includegraphics[width=0.8\linewidth]{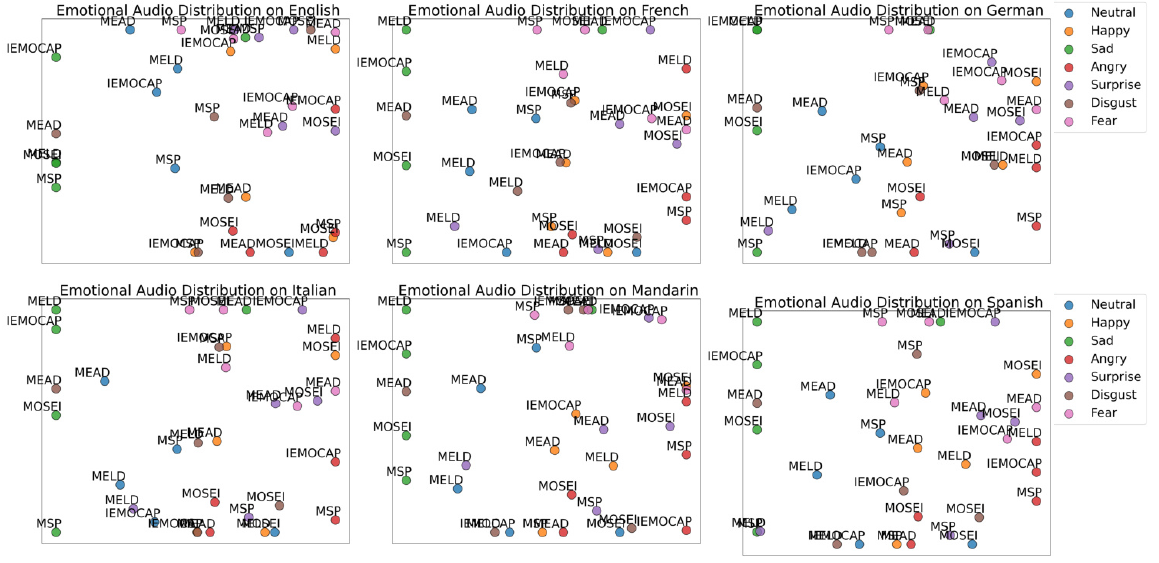}}
\vspace{-0.3cm}
\caption{Comparison of emotional audio distribution among English and languages in M5SER.}
\label{fig025}
\vspace{-0.3cm}
\end{figure*}

\section{M5SER: Multilingual SER Dataset}
\label{sec_dataset}
In this section, we introduce the synthetic multilingual Speech Emotion Recognition (SER) dataset, M5SER, which comprises emotional audio recordings in five languages.

\subsection{Overview of M5SER}
M5SER is a large-scale multilingual speech emotion dataset, consisting of emotional audio signals in five languages, covering seven basic emotions: neutral, happy, sad, angry, surprise, disgust, and fear. In total, it contains over 1009 hours of generated emotional audio signals. As shown in Fig. \ref{fig02}, the audio signals in various languages are balanced across both the generated M5SER dataset and raw English audios.

\subsection{Construction of M5SER}

To form a large-scale speech emotion dataset, first, we gather data with human-annotated emotion labels from existing English-based SER datasets, including IEMOCAP \cite{busso2008iemocap}, MELD \cite{poria2019meld}, MAED \cite{wang2020mead}, CMU-MOSEI \cite{zadeh2018multimodal}, and MSP-Podcast \cite{lotfian2017building}. These emotional audio signals serve as source to synthesize emotion-preserving audios in various languages.

\emph{Data Generation}: In the second step, we employ the foundation model SeamlessExpressive \cite{barrault2023seamless} as an audio translator to generate new audio signals. It comprises of two main modules: Prosody UnitY2, a prosody-aware speech-to-unit translation model based on the UnitY2 architecture, and PRETSSEL, a unit-to-speech model that preserves cross-lingual expressivity. These modules excel in emotion-aware speech translation. We use the emotion-annotated audio signals as the speech source to generate new speech in five additional languages: French, German, Italian, Mandarin, and Spanish.

\emph{Data Filtering}: In the third step, we sample and remove low-quality generated audio signal using two criteria: duration and size. Audios with duration shorter than 0.5 seconds are removed. Due to the typically smaller size of audios in IEMOCAP and MELD, we remove audio signals from these datasets if they are smaller than 20KB. For MAED, MOSEI, and MSP, we remove audios smaller than 50KB.

\subsection{Analysis of M5SER}
To evaluate the consistency of emotion-aware speech translation, we analyze the mean spectrograms of audio samples representing various emotions across multiple datasets and languages. If the generated audio samples for the same emotion across different languages cluster closely in the spectrogram mean space, it indicates that the multilingual audio effectively retains the emotional distribution of the original speech. This is based on the widely accepted premise that spectrogram features are reliable indicators of emotional content in speech \cite{koolagudi2012emotion}. As illustrated in Fig. \ref{fig025}, we visualize the distributions of the mean spectrograms from both the generated M5SER dataset and the original English-based SER dataset. The results show that the mean spectrograms of audio samples with the same emotion exhibit consistent distributions across different languages and datasets. This demonstrates the high quality and emotion-preserving capability of our generated emotion-aware multilingual speech emotion dataset.

\begin{figure*}[htbp]
\vspace{-0.5cm}
\centerline{\includegraphics[width=0.8\linewidth]{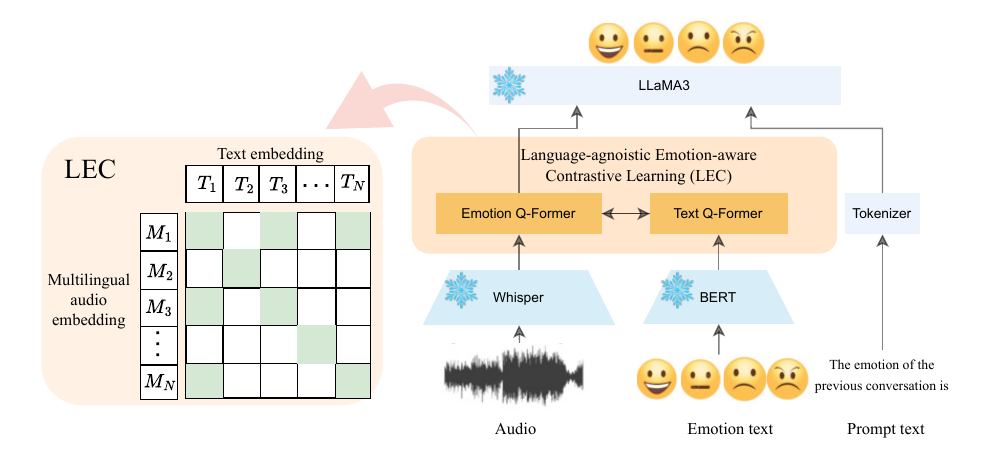}}
\vspace{-0.3cm}
\caption{Overview of multilingual speech emotion recognition framework.}
\label{fig01}
\vspace{-0.3cm}
\end{figure*}

\section{Method}

Given the success of audio understanding tasks in existing audio-conditioned large language models (LLMs), we formulate the zero-shot MSER task as a zero-shot speech emotion word generation task in this work.

\subsection{Model Architecture}

Following the design of audio-integrated LLMs for general audio understanding, we employ an encoder-decoder architecture. As shown in Fig. \ref{fig01}, we utilize an audio encoder, an LLM decoder, and a cross-modal connector that facilitates the alignment between audio and language representations.

\noindent\textbf{Audio Encoder.} As shown in Fig. \ref{fig01}, we employ the Whisper model as the audio encoder to extract speech features from multilingual audio sources. It is pretrained with multitasking learning on multilingual resources and can be used for multilingual speech recognition, speech translation, and language identification. Following previous works, the parameters of the audio encoder are frozen to avoid heavy computational costs.

\noindent\textbf{Modality Interface.} The extracted features are then passed through a cross-modal connector, which aligns the audio representations with the language model's latent space, enabling seamless integration of audio and text modalities for comprehensive audio-language understanding. To achieve both cross-modal alignment and audio feature compression, we employ the Emotion Q-Former as the connector. The parameters of the Emotion Q-Former are initialized from BERT, incorporating self-attention, cross-attention, and linear layers. Learnable query embeddings are used to extract emotion-aware speech representations, enhancing the model's capacity to capture emotional nuances from the speech input.

\noindent\textbf{Large Language Models.} Finally, we utilize LLaMA 3, a state-of-the-art LLM, as the emotion estimator. It is used to generate the corresponding emotion words from the audio input. Considering the input format of LLaMA 3, we set our input as [bos, audio, prompt, [emo]], starting with sign of bos.

\subsection{Multilingual Speech Representation Learning}

We propose to utilize contrastive learning to learn multilingual emotion-aware speech representations to endow LLMs for zero-shot MSER. This process contains two stages: emotion-aware speech representation learning and multilingual emotion-aware speech representation learning.

\subsubsection{Emotion-Aware Speech Representation Learning}

To address the high dimensionality of encoded speech representations, which often include irrelevant information like background noise, we introduce the Emotion Q-Former to extract emotion-specific features while compressing the speech data. We leverage an emotion-aware contrastive learning approach combined with the Emotion Q-Former to enhance the LLM decoder's performance in speech emotion decoding. This stage of training relies on human-annotated English speech emotion datasets, which include various speakers and speech contents and are used to learn emotion-aware, speaker-independent speech representations through a contrastive design.

To mitigate the impact of speaker and content variability in audio signals, we use emotion labels as the primary discriminative factor during training. Cosine similarity is employed to quantify the distance between speech features. Audio samples sharing the same emotion label are treated as positive pairs, with the objective of maximizing the similarity between their speech representations. Conversely, audio samples with differing emotion labels are considered negative pairs, and we aim to minimize the similarity between their speech representations.

\vspace{-0.2cm}
\begin{equation} 
\small 
\vspace{-0.2cm}
\begin{aligned} 
\mathcal{L}_{lec} = & \sum_{i=1}^{b} \left[ \sum_{j=1}^{b} \left( w_1 y_{ij} (1 - S) \right. \right. \\ 
& \quad + \left. \left. w_2 (1 - y_{ij}) \max(0, (S - m)) \right) \right] 
\end{aligned} 
\label{eq:eq01} 
\end{equation}
\noindent where $b$ is the batch size, $S$ is the similarity between audio utterances,  $y_{ij}$ is the result of the different or different labels of sample $i$ and $j$ and $m$ is the margin.

\subsubsection{Emotion-Aware Multilingual Speech Enhancement}

Large-scale multilingual speech emotion datasets are crucial for capturing cross-lingual and language-agnostic emotion-aware speech representations, yet they remain absent in the existing MSER field. In this paper, we introduce a large-scale multilingual speech emotion dataset and use it for multilingual speech emotion recognition learning. Compared to human-annotated datasets, those generated by large foundation models can be much larger at a fraction of the cost. These datasets are increasingly used to train multimodal large language models, and we are the first to introduce large-scale synthetic speech data for MSER training.

\subsubsection{Language-Agnostic Emotion-Aware Speech Representation Learning}
Following the stage of emotion-aware speech representation learning, we also apply contrastive loss during the emotion words decoding process.  The difference in the contrastive design between the second stage and the first stage is the variety of data resources. Compared with the human-annotated English SER datasets in the first stage, the synthetic multilingual SER datasets are used in the second stage. Audio with different languages help to capture the language-agnostic emotion-aware speech information under the contrastive design.

\subsection{Training Process}
As mentioned above, we introduce two stages of training to contrastively align speech signals with language features for zero-shot MSER. To save the computing costs, we freeze both the audio encoder and the LLM during these two stages. In the first stage, we combine the emotion-aware contrastive (EC) loss with the LLM decoding loss for collaborative training. Specifically, the training loss is 

\vspace{-0.2cm}
\begin{equation} \small
\vspace{-0.2cm}
    \begin{aligned}
        \mathcal{L}_{s1} = \mathcal{L}_{lec} + \lambda \cdot \mathcal{L}_{ce}
        \label{eq:eq04}
    \end{aligned}
\end{equation}

In the second stage, we combine the language-agnostic emotion-aware contrastive (LEC) with the LLM decoding loss for collaborative training. Specifically, the training loss is
\vspace{-0.2cm}
\begin{equation} \small
\vspace{-0.3cm}
    \begin{aligned}
        \mathcal{L}_{s2} =  \mathcal{L}_{lec} +   \lambda  \cdot \mathcal{L}_{ce}
        \label{eq:eq05}
    \end{aligned}
\end{equation}

\begin{table*}[htbp] \small
\vspace{-0.5cm}
\caption{Main results of {\color{blue} \textbf{a)}} 
 traditional speech emotion recognition and {\color{blue} \textbf{b)}} zero-shot multilingual speech emotion recognition.}
\label{tab01}
\begin{center}
\setlength\tabcolsep{4pt}
\vspace{-0.2cm}
\scalebox{0.95}{\begin{tabular}{lccccccccc} 
    \toprule
    \multirow{2}*{{\color{blue} \textbf{a)}}  \textbf{Model}}  &  \multicolumn{4}{c}{\textbf{IEMOCAP}} & & \multicolumn{4}{c}{\textbf{MELD}}    \\ 
    \cmidrule{2-5}  \cmidrule{7-10}
     & \#{E} & WA & UA & WF1 & & \#{E} & WA & UA & WF1   \\
    \midrule
    MCED \cite{li2021speech} & 4 & -- & 61.4 & 60.1 &  & 7 & 49.7 & -- &44.6\\
    MTL-MLi \cite{sharma2022multi} & 7 & -- & -- & 64.1 &   & 7 & -- & -- & 49.8\\
    ASER \cite{feng2024foundation} & 4 & -- & 67.45 & -- & & 4 & -- &  48.55 & -- \\
    Vesper \cite{chen2024vesper} & 4 & 70.7 & 70.8 & 70.6 & & 7 & 53.5 & 26.8 & 48.0 \\
    SpeechFormer++ \cite{chen2023speechformer++} & 4 & 70.5 & 71.5 & 70.7 &  & 7 & 51.0 & 27.3 & 47.0 \\
    emotion2vec \cite{ma2023emotion2vec} & 4 & 71.8 & -- & -- &  & 7 & 51.9 & 28.0 & 48.7 \\
    \midrule
    \textbf{Ours} & 4   & 70.3 & 72.4 & 70.3 &  & 4 & 61.4 & 52.6 & 59.9     \\ 
    \textbf{Ours} & 7   & 69.0 & 51.3 & 68.5 &  & 7 & 54.9 & 31.4  & 49.8    \\ 
    \bottomrule
\end{tabular}}
\quad
\scalebox{0.9}{\begin{tabular}{lcccc}
    \toprule
    \multirow{2}*{{\color{blue} \textbf{a)}}  \textbf{Dataset}}  & \multirow{2}*{  \textbf{Model}} &  \multicolumn{3}{c}{\textbf{Performance}}  \\ 
    \cmidrule{3-5} 
     & & WA & UA & WF1  \\
    \midrule  
    \multirow{2}*{ShEMO (6)} &  SOTA (PreTrain+Train) & 80.0 &  66.0 & 79.6  \\
    & \textbf{Ours ZS} & 71.3 & 57.4 & 71.8  \\
    \midrule  
    \multirow{2}*{AESDD(5)} & SOTA (PreTrain+Train) & 72.3 & 72.3 & 71.6 \\
    &  \textbf{Ours ZS} & 42.6 & 42.6 & 36.5 \\
    \midrule  
    \multirow{2}*{RAVDESS (7)} & SOTA (PreTrain+Train) & 82.4 & 82.9 & 82.4 \\
    & \textbf{Ours ZS} & 41.7 & 45.3 & 34.8 \\
    \midrule   
    \multirow{2}*{RESD (7)} &  SOTA (PreTrain+Train) & 64.8 & 65.0 & 64.5 \\
    & \textbf{Ours ZS} &  39.2 & 33.0 & 33.2  \\
    \bottomrule
\end{tabular}}  
\end{center}
\end{table*}

\begin{table}[htbp] \small
\vspace{-0.5cm}
\caption{Ablation study of speech emotion recognition on IEMOCAP.}
\label{tab02}
\begin{center}
\setlength\tabcolsep{10pt}
\scalebox{0.9}{\begin{tabular}{lcccc} 
    \toprule
    \multirow{2}*{{\color{blue} \textbf{}}  \textbf{Model}}  &  \multicolumn{4}{c}{\textbf{IEMOCAP}} \\ 
    \cmidrule{2-5} 
     &  WA & UA & WF1 & Precision   \\
    \midrule
    SER & 59.7 & 48.8 & 58.4 & 67.9\\
    SER+$\mathcal{L}_{lec}$ & 64.7 & 48.6 & 64.0 & 67.9\\
    \midrule
    MSER  & 62.2 & 50.0 & 61.1 & 70.1 \\
    MSER+$\mathcal{L}_{lec}$ & 68.6 & 49.1 & 67.8 & 70.2 \\
    \midrule
    MSER (S2) & 66.0 & 49.5 & 65.0 & 71.0 \\
    MSER+$\mathcal{L}_{lec}$ (S2) & 69.0 & 51.3 & 68.5 & 71.0\\
    \bottomrule
\end{tabular}}
\end{center}
\vspace{-0.2cm}
\end{table}

\section{Experiments}
In this section, we present the datasets and setup used for our training and evaluation process. We also outline the baseline models and evaluation metrics employed in the subsequent section for performance comparison and results analysis.

\subsection{Datasets}
For the first-stage training, we utilize the previously mentioned human-annotated English SER datasets (see Sec. \ref{sec_dataset}) and our synthetic multilingual dataset, M5SER. In addition to basic SER evaluation on the unseen samples of the IEMOCAP and MELD, we also focus on zero-shot MSER evaluation. The extra unseen datasets used for zero-shot MSER evaluation include AESDD \cite{vryzas2018speech} (Greek, 5 emotions), RAVDESS \cite{livingstone2018ryerson} (English, 7 emotions), RESD \cite{Aniemore} (Russian, 7 emotions), and ShEMO \cite{mohamad2019shemo} (Persian, 6 emotions).

\subsection{Experimental Settings}
For the multilingual speech emotion recognition model, we fine-tune only the parameters of the modality connectors, keeping the parameters of both the modality encoders and the LLM fixed. We adopt the AdamW optimization algorithm with a learning rate of $1 \times 10^{-5}$, and to mitigate overfitting, we employ a weight decay of $1 \times 10^{-6}$. The Q-Former structure uses a query embedding size of 256, regardless of whether the parameters are initialized from BERT. To compare performance, we employ commonly used metrics including weighted accuracy (WA), unweighted accuracy (UA), weighted F1 score (WF1), and precision. Higher values for these metrics indicate better model performance.

\section{Results and Analysis}

\subsection{Main results}
In Table \ref{tab01} (a), our approach achieves a WA of 70.3\% and 68.5\% for four-class and seven-class tasks on the IEMOCAP dataset, respectively, which is comparable to the best-performing method (71.8\% WA for four-class and 70.8\% WA for seven-class). On the MELD dataset, our method achieves a WA of 61.4\% for four-class and 54.9\% for seven-class tasks, demonstrating competitive performance. In Table \ref{tab01} (b), our zero-shot MSER method also shows competitive performance with the current SOTA SER method that is pretrained on multiple SER sources and then fine-tuned on the corresponding dataset \cite{ma2023emotion2vec}. This demonstrates the effectiveness and generality of our approach on unseen datasets and even unseen languages.

\begin{figure}[htbp]
\vspace{-0.5cm}
\centerline{\includegraphics[width=1.0\linewidth]{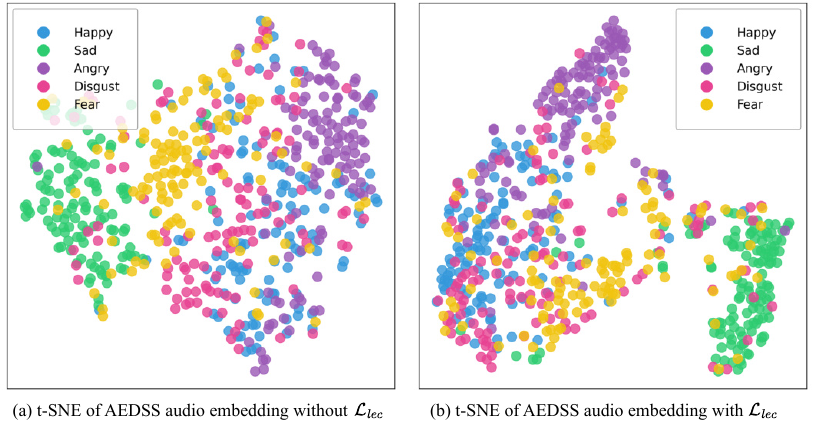}}
\vspace{-0.2cm}
\caption{Audio distribution with t-SNE embedding.}
\label{fig04}
\end{figure}

\subsection{Ablation Study}
The ablation study in Table \ref{tab02} systematically evaluates the contributions of the proposed multilingual dataset and contrastive loss. Training solely on the original English dataset achieves a WA of 59.7\%, while incorporating $L_{ec}$ improves WA to 64.7\%. When trained on the multilingual dataset, performance increases significantly, achieving a WA of 62.2\%, further enhanced to 68.6\% with $L_{lec}$. Notably, the two-stage training strategy, initial training on English data followed by multilingual fine-tuning, achieves a WA of 66.0\%, further boosted to 69.0\% when $L_{lec}$. These results validate the effectiveness of the multilingual dataset and contrastive loss, as well as the two-stage training approach, in improving model performance on speech emotion recognition tasks.

\subsection{Analysis}
The t-SNE plots in Fig. \ref{fig04} illustrate the benefits of incorporating the language-agnostic emotion-aware contrastive loss \(L_{lec}\) in the audio embeddings of the AEDSS dataset. In plot (a), without the contrastive loss, the emotion categories are more dispersed and intermixed, indicating less distinct separation between different emotions. In contrast, plot (b) shows the audio embeddings with the contrastive loss \(L_{ec}\), where the emotion categories are more clearly clustered and separated. This demonstrates that the contrastive loss helps in better distinguishing between different emotion categories, leading to more distinct and meaningful embeddings.

\section{Conclusion}
In this paper, we propose a novel framework for zero-shot multilingual speech emotion recognition (MSER), leveraging contrastive alignment between multilingual speech and emotion-aware language representations. By introducing a large-scale synthetic MSER dataset and integrating a multilingual audio encoder with a LLM, our method achieves state-of-the-art performance on both commonly SER and the zero-shot MSER tasks on unseen datasets and languages. These results highlight the framework's ability to generalize across languages and datasets, advancing the field of multi-lingual speech emotion recognition.

\section*{Acknowledgment}
This work has been supported by the National Research Foundation Singapore under AI Singapore Programme (Award Number: AISG2-TC-2022-004).

\bibliographystyle{IEEEtran}
\bibliography{icme2025references}

\begin{thebibliography}{10}
\providecommand{\url}[1]{#1}
\csname url@samestyle\endcsname
\providecommand{\newblock}{\relax}
\providecommand{\bibinfo}[2]{#2}
\providecommand{\BIBentrySTDinterwordspacing}{\spaceskip=0pt\relax}
\providecommand{\BIBentryALTinterwordstretchfactor}{4}
\providecommand{\BIBentryALTinterwordspacing}{\spaceskip=\fontdimen2\font plus
\BIBentryALTinterwordstretchfactor\fontdimen3\font minus \fontdimen4\font\relax}
\providecommand{\BIBforeignlanguage}[2]{{%
\expandafter\ifx\csname l@#1\endcsname\relax
\typeout{** WARNING: IEEEtran.bst: No hyphenation pattern has been}%
\typeout{** loaded for the language `#1'. Using the pattern for}%
\typeout{** the default language instead.}%
\else
\language=\csname l@#1\endcsname
\fi
#2}}
\providecommand{\BIBdecl}{\relax}
\BIBdecl

\bibitem{busso2008iemocap}
C.~Busso, M.~Bulut, C.-C. Lee, A.~Kazemzadeh, E.~Mower, S.~Kim, J.~N. Chang, S.~Lee, and S.~S. Narayanan, ``Iemocap: Interactive emotional dyadic motion capture database,'' \emph{Language resources and evaluation}, vol.~42, pp. 335--359, 2008.

\bibitem{livingstone2018ryerson}
S.~R. Livingstone and F.~A. Russo, ``The ryerson audio-visual database of emotional speech and song (ravdess): A dynamic, multimodal set of facial and vocal expressions in north american english,'' \emph{PloS one}, vol.~13, no.~5, p. e0196391, 2018.

\bibitem{chen2024vesper}
W.~Chen, X.~Xing, P.~Chen, and X.~Xu, ``Vesper: A compact and effective pretrained model for speech emotion recognition,'' \emph{IEEE Transactions on Affective Computing}, 2024.

\bibitem{al2022transformer}
B.~B. Al-onazi, M.~A. Nauman, R.~Jahangir, M.~M. Malik, E.~H. Alkhammash, and A.~M. Elshewey, ``Transformer-based multilingual speech emotion recognition using data augmentation and feature fusion,'' \emph{Applied Sciences}, vol.~12, no.~18, p. 9188, 2022.

\bibitem{zehra2021cross}
W.~Zehra, A.~R. Javed, Z.~Jalil, H.~U. Khan, and T.~R. Gadekallu, ``Cross corpus multi-lingual speech emotion recognition using ensemble learning,'' \emph{Complex \& Intelligent Systems}, vol.~7, no.~4, pp. 1845--1854, 2021.

\bibitem{sharma2022multi}
M.~Sharma, ``Multi-lingual multi-task speech emotion recognition using wav2vec 2.0,'' in \emph{ICASSP 2022-2022 IEEE International Conference on Acoustics, Speech and Signal Processing (ICASSP)}.\hskip 1em plus 0.5em minus 0.4em\relax IEEE, 2022, pp. 6907--6911.

\bibitem{ma2024emobox}
Z.~Ma, M.~Chen, H.~Zhang, Z.~Zheng, W.~Chen, X.~Li, J.~Ye, X.~Chen, and T.~Hain, ``Emobox: Multilingual multi-corpus speech emotion recognition toolkit and benchmark,'' \emph{arXiv preprint arXiv:2406.07162}, 2024.

\bibitem{li2023zero}
Y.~Li, X.~Zhu, Y.~Lei, H.~Li, J.~Liu, D.~Xie, and L.~Xie, ``Zero-shot emotion transfer for cross-lingual speech synthesis,'' in \emph{2023 IEEE Automatic Speech Recognition and Understanding Workshop (ASRU)}.\hskip 1em plus 0.5em minus 0.4em\relax IEEE, 2023, pp. 1--8.

\bibitem{gao2023adversarial}
Y.~Gao, L.~Wang, J.~Liu, J.~Dang, and S.~Okada, ``Adversarial domain generalized transformer for cross-corpus speech emotion recognition,'' \emph{IEEE Transactions on Affective Computing}, 2023.

\bibitem{lian2024exploring}
H.~Lian, C.~Lu, Y.~Zhao, S.~Li, T.~Qi, and Y.~Zong, ``Exploring corpus-invariant emotional acoustic feature for cross-corpus speech emotion recognition,'' \emph{Expert Systems with Applications}, vol. 258, p. 125162, 2024.

\bibitem{zhao2024emotion}
Y.~Zhao, J.~Wang, C.~Lu, S.~Li, B.~W. Schuller, Y.~Zong, and W.~Zheng, ``Emotion-aware contrastive adaptation network for source-free cross-corpus speech emotion recognition,'' in \emph{ICASSP 2024-2024 IEEE International Conference on Acoustics, Speech and Signal Processing (ICASSP)}.\hskip 1em plus 0.5em minus 0.4em\relax IEEE, 2024, pp. 11\,846--11\,850.

\bibitem{gomez2024speech}
L.~G{\'o}mez-Zaragoz{\'a}, {\'O}.~Valls, R.~del Amor, M.~J. Castro-Bleda, V.~Naranjo, M.~A. Raya, and J.~Mar{\'\i}n-Morales, ``Speech emotion recognition from voice messages recorded in the wild,'' \emph{arXiv preprint arXiv:2403.02167}, 2024.

\bibitem{chu2023qwen}
Y.~Chu, J.~Xu, X.~Zhou, Q.~Yang, S.~Zhang, Z.~Yan, C.~Zhou, and J.~Zhou, ``Qwen-audio: Advancing universal audio understanding via unified large-scale audio-language models,'' \emph{arXiv preprint arXiv:2311.07919}, 2023.

\bibitem{radford2023robust}
A.~Radford, J.~W. Kim, T.~Xu, G.~Brockman, C.~McLeavey, and I.~Sutskever, ``Robust speech recognition via large-scale weak supervision,'' in \emph{International conference on machine learning}.\hskip 1em plus 0.5em minus 0.4em\relax PMLR, 2023, pp. 28\,492--28\,518.

\bibitem{santoso2024large}
J.~Santoso, K.~Ishizuka, and T.~Hashimoto, ``Large language model-based emotional speech annotation using context and acoustic feature for speech emotion recognition,'' in \emph{ICASSP 2024-2024 IEEE International Conference on Acoustics, Speech and Signal Processing (ICASSP)}.\hskip 1em plus 0.5em minus 0.4em\relax IEEE, 2024, pp. 11\,026--11\,030.

\bibitem{xu2024secap}
Y.~Xu, H.~Chen, J.~Yu, Q.~Huang, Z.~Wu, S.-X. Zhang, G.~Li, Y.~Luo, and R.~Gu, ``Secap: Speech emotion captioning with large language model,'' in \emph{Proceedings of the AAAI Conference on Artificial Intelligence}, vol.~38, no.~17, 2024, pp. 19\,323--19\,331.

\bibitem{bukhari2024selm}
H.~Bukhari, S.~Deshmukh, H.~Dhamyal, B.~Raj, and R.~Singh, ``Selm: Enhancing speech emotion recognition for out-of-domain scenarios,'' \emph{arXiv preprint arXiv:2407.15300}, 2024.

\bibitem{zou2022speech}
H.~Zou, Y.~Si, C.~Chen, D.~Rajan, and E.~S. Chng, ``Speech emotion recognition with co-attention based multi-level acoustic information,'' in \emph{ICASSP 2022-2022 IEEE International Conference on Acoustics, Speech and Signal Processing (ICASSP)}.\hskip 1em plus 0.5em minus 0.4em\relax IEEE, 2022, pp. 7367--7371.

\bibitem{floridi2020gpt}
L.~Floridi and M.~Chiriatti, ``Gpt-3: Its nature, scope, limits, and consequences,'' \emph{Minds and Machines}, vol.~30, pp. 681--694, 2020.

\bibitem{ericsson2022self}
L.~Ericsson, H.~Gouk, C.~C. Loy, and T.~M. Hospedales, ``Self-supervised representation learning: Introduction, advances, and challenges,'' \emph{IEEE Signal Processing Magazine}, vol.~39, no.~3, pp. 42--62, 2022.

\bibitem{zou2023unis}
H.~Zou, M.~Shen, C.~Chen, Y.~Hu, D.~Rajan, and E.~S. Chng, ``Unis-mmc: Multimodal classification via unimodality-supervised multimodal contrastive learning,'' \emph{arXiv preprint arXiv:2305.09299}, 2023.

\bibitem{chen2020simple}
T.~Chen, S.~Kornblith, M.~Norouzi, and G.~Hinton, ``A simple framework for contrastive learning of visual representations,'' in \emph{International conference on machine learning}.\hskip 1em plus 0.5em minus 0.4em\relax PMLR, 2020, pp. 1597--1607.

\bibitem{zou2024cross}
H.~Zou, M.~Shen, Y.~Hu, C.~Chen, E.~S. Chng, and D.~Rajan, ``Cross-modality and within-modality regularization for audio-visual deepfake detection,'' in \emph{ICASSP 2024-2024 IEEE International Conference on Acoustics, Speech and Signal Processing (ICASSP)}.\hskip 1em plus 0.5em minus 0.4em\relax IEEE, 2024, pp. 4900--4904.

\bibitem{poria2019meld}
S.~Poria, D.~Hazarika, N.~Majumder, G.~Naik, E.~Cambria, and R.~Mihalcea, ``Meld: A multimodal multi-party dataset for emotion recognition in conversations,'' in \emph{Proceedings of the 57th Annual Meeting of the Association for Computational Linguistics}, 2019, pp. 527--536.

\bibitem{wang2020mead}
K.~Wang, Q.~Wu, L.~Song, Z.~Yang, W.~Wu, C.~Qian, R.~He, Y.~Qiao, and C.~C. Loy, ``Mead: A large-scale audio-visual dataset for emotional talking-face generation,'' in \emph{European Conference on Computer Vision}.\hskip 1em plus 0.5em minus 0.4em\relax Springer, 2020, pp. 700--717.

\bibitem{zadeh2018multimodal}
A.~B. Zadeh, P.~P. Liang, S.~Poria, E.~Cambria, and L.-P. Morency, ``Multimodal language analysis in the wild: Cmu-mosei dataset and interpretable dynamic fusion graph,'' in \emph{Proceedings of the 56th Annual Meeting of the Association for Computational Linguistics (Volume 1: Long Papers)}, 2018, pp. 2236--2246.

\bibitem{lotfian2017building}
R.~Lotfian and C.~Busso, ``Building naturalistic emotionally balanced speech corpus by retrieving emotional speech from existing podcast recordings,'' \emph{IEEE Transactions on Affective Computing}, vol.~10, no.~4, pp. 471--483, 2017.

\bibitem{barrault2023seamless}
L.~Barrault, Y.-A. Chung, M.~C. Meglioli, D.~Dale, N.~Dong, M.~Duppenthaler, P.-A. Duquenne, B.~Ellis, H.~Elsahar, J.~Haaheim \emph{et~al.}, ``Seamless: Multilingual expressive and streaming speech translation,'' \emph{arXiv preprint arXiv:2312.05187}, 2023.

\bibitem{koolagudi2012emotion}
S.~G. Koolagudi and K.~S. Rao, ``Emotion recognition from speech: a review,'' \emph{International journal of speech technology}, vol.~15, pp. 99--117, 2012.

\bibitem{li2021speech}
R.~Li, J.~Zhao, and Q.~Jin, ``Speech emotion recognition via multi-level cross-modal distillation.'' in \emph{Interspeech}, 2021, pp. 4488--4492.

\bibitem{feng2024foundation}
T.~Feng and S.~Narayanan, ``Foundation model assisted automatic speech emotion recognition: Transcribing, annotating, and augmenting,'' in \emph{ICASSP 2024-2024 IEEE International Conference on Acoustics, Speech and Signal Processing (ICASSP)}.\hskip 1em plus 0.5em minus 0.4em\relax IEEE, 2024, pp. 12\,116--12\,120.

\bibitem{chen2023speechformer++}
W.~Chen, X.~Xing, X.~Xu, J.~Pang, and L.~Du, ``Speechformer++: A hierarchical efficient framework for paralinguistic speech processing,'' \emph{IEEE/ACM Transactions on Audio, Speech, and Language Processing}, vol.~31, pp. 775--788, 2023.

\bibitem{ma2023emotion2vec}
Z.~Ma, Z.~Zheng, J.~Ye, J.~Li, Z.~Gao, S.~Zhang, and X.~Chen, ``emotion2vec: Self-supervised pre-training for speech emotion representation,'' \emph{arXiv preprint arXiv:2312.15185}, 2023.

\bibitem{vryzas2018speech}
N.~Vryzas, R.~Kotsakis, A.~Liatsou, C.~A. Dimoulas, and G.~Kalliris, ``Speech emotion recognition for performance interaction,'' \emph{Journal of the Audio Engineering Society}, vol.~66, no.~6, pp. 457--467, 2018.

\bibitem{Aniemore}
N.~D. Artem~Amentes and I.~Lubenets, ``Resd (russian emotional speech dialogs with annotated text),'' 2022.

\bibitem{mohamad2019shemo}
O.~Mohamad~Nezami, P.~Jamshid~Lou, and M.~Karami, ``Shemo: a large-scale validated database for persian speech emotion detection,'' \emph{Language Resources and Evaluation}, vol.~53, pp. 1--16, 2019.

\end{thebibliography}

\end{document}